%% file: manuscript.tex
\documentclass[a4paper]{article}
\pdfoutput=1

\usepackage{IEK10} 
\usepackage{natbib}

\def\IEK10{
  Forschungszentrum Jülich GmbH,
  Institute of Energy and Climate Research,
  Energy Systems Engineering (IEK-10),
  Jülich 52425,
  Germany
}
\def\RWTH{
  RWTH Aachen University
  Aachen 52062,
  Germany
}
\def\JARA{
  JARA-ENERGY,
  Jülich 52425,
  Germany
}
\def\SVT{
  RWTH Aachen University,
  Process Systems Engineering (AVT.SVT),
  Aachen 52074,
  Germany
}
\def\UQ{
  RWTH Aachen University,
  Institute of Uncertainty Quantification,
  Aachen 52062,
  Germany
}
\def\KAUST{
  King Abdullah University of Science and Technology (KAUST),
  Computer, Electrical, and Mathematical Sciences \& Engineering Division (CEMSE),
  Saudi Arabia
}

\newcommand{\mytitle}{Principal Component Density Estimation for Scenario Generation Using Normalizing Flows}

\newcommand{\affil}{
  \begin{itemize}[leftmargin=3mm, itemsep=0mm]
    \item[$^a$]\IEK10
    \item[$^b$]\RWTH
    \item[$^c$]\JARA
    \item[$^d$]\SVT
    \item[$^e$]\UQ
    \item[$^f$]\KAUST
  \end{itemize}
}

\def\firstAuthor{Eike Cramer}
\newcommand{\myauthor}{\firstAuthor$^{a,b}$, Alexander Mitsos$^{c,a,d}$, Ra\'ul Tempone$^{e,f}$, Manuel Dahmen$^{a,*}$ }
\newcommand{\correspondingauthor}{Manuel Dahmen, \IEK10 \newline E-mail: m.dahmen@fz-juelich.de}
\author{\myauthor}

\usepackage[
  colorlinks,
  linkcolor=blue,
  citecolor=blue,
  urlcolor=blue,
  pdftitle={\mytitle},
  pdfauthor={\firstAuthor}
]{hyperref}
\usepackage[capitalise, nameinlink]{cleveref}
\crefname{table}{Tab.}{Tab.}

\begin{document}

  \thispagestyle{firststyle}

  \begin{center}
    \begin{large}
      \textbf{\mytitle}
    \end{large} \\
    \myauthor
  \end{center}

  \vspace{0.5cm}

  \begin{footnotesize}
    \affil
  \end{footnotesize}

  \vspace{0.5cm}

  \begin{abstract}
    \input{sections/00_Abstract.tex}
  \end{abstract}
    
  \vspace{0.5cm}
  

  \input{sections/01_Introduction.tex}
  \input{sections/02_NormalizingFlows.tex}

  \input{sections/03_PrincipalComponentFlow.tex}
  \input{sections/04_NumericalExperiments.tex}
  \input{sections/05_Conclusion.tex}
  \input{sections/acknowledgement.tex}
  \input{sections/appendices.tex}

  \bibliographystyle{apalike}
  \renewcommand{\refname}{Bibliography}
  \bibliography{literature.bib}

\end{document}

%% file: sections/00_Abstract.tex
Neural networks-based learning of the distribution of non-dispatchable renewable electricity generation from sources such as photovoltaics (PV) and wind as well as load demands has recently gained attention.
Normalizing flow density models are particularly well suited for this task due to the training through direct log-likelihood maximization.
However, research from the field of image generation has shown that standard normalizing flows can only learn smeared-out versions of manifold distributions. 
Previous works on normalizing flow-based scenario generation do not address this issue, and the smeared-out distributions result in the sampling of noisy time series.
In this paper, we exploit the isometry of the principal component analysis (PCA), which sets up the normalizing flow in a lower-dimensional space while maintaining the direct and computationally efficient likelihood maximization. 
We train the resulting principal component flow (PCF) on data of PV and wind power generation as well as load demand in Germany in the years 2013 to 2015.
The results of this investigation show that the PCF preserves critical features of the original distributions, such as the probability density and frequency behavior of the time series.
The application of the PCF is, however, not limited to renewable power generation but rather extends to any data set, time series, or otherwise, which can be efficiently reduced using PCA.

%% file: sections/01_Introduction.tex
\section{Introduction}\label{sec:Introduction}
The renewable electricity generation technologies photovoltaics (PV) and wind depend on natural occurrences and are therefore non-dispatchable.
Additionally, the realization of these renewable power generation outputs exhibits uncertain and volatile behavior, which poses new challenges for the design and operation of energy systems compared to dispatchable fossil power generation \citep{AgoraEnergiewende2020,mitsos2018challenges}. 
To account for the uncertainty in renewable electricity sources and other relevant energy system parameters like electricity demands system operators require information about possible realizations. This information can either be provided via short-term point-forecasting \citep{tascikaraoglu2014review} or via consideration of multiple scenarios \citep{morales2013integrating}. Here, the term scenario refers to a possible realization of the uncertain and volatile parameter over a certain time span. 
A set of scenarios thus refers to a collection of time series of equal length. Decision-making based on scenarios covers the broad spectrum of possible realizations and is, therefore, highly relevant for design problems \citep{morales2013integrating}. 
Scenarios can be applied in stochastic programming formulations for the design and operation of energy systems as well as energy-related scheduling tasks \citep{kaut2003evaluation,birge2011introduction}.  
The distributions of time series intervals are often unknown and consist of many non-independent dimensions due to the correlation between time steps. These distributions typically do not follow standard distribution models like multivariate Gaussians. Thus, sampling these distributions for scenario generation remains an open research question. 

There are many contributions to the literature on how to generate scenarios. In general, the published approaches can be classified as univariate modeling approaches, i.e., step-by-step models, and multivariate modeling approaches, where multiple time steps are modeled in parallel \citep{ziel2018day}. 
Multivariate modeling approaches are trained on sets of scenarios created from historical time series. The equidistant segments of univariate scenarios are viewed as multidimensional points, e.g., a scenario of one day with hourly recordings is viewed as a 24-dimensional data point. 
Contributions on univariate modeling approaches include the traditional Box-Jenkins approach \citep{box1970time} for sampling stochastic processes \citep{sharma2013wind} and artificial neural network (ANN)-based autoregressive models \citep{vagropoulos2016ann}. 
Examples of multivariate modeling approaches are Gaussian mixture models \citep{wang2018conditional}, Copula methods \citep{pinson2009probabilistic,kaut2011shape}, and moment matching techniques \citep{chopra2020scenario}. 

With the increase of computational power and the development of advanced machine learning algorithms, it is now possible to train specialized ANNs, so-called deep generative models (DGMs), to learn high-dimensional probability distributions without any statistical assumptions about the data. Thus, DGMs are powerful tools for multivariate modeling.
In 2018, \cite{chen2018model} proposed the use of generative adversarial networks (GANs) \citep{goodfellow2014generative} for scenario generation. Since then, scenario generation using GANs and other DGMs has become a popular topic \citep{chen2018model,jiang2018scenario,jiang2019forecasting,zhang2020typical,wei2019short,zhang2019scenario}. 
\cite{chen2018model} later extended their original work by introducing a scenario selection procedure \citep{chen2018unsupervised} and by using Bayesian GANs \citep{chen2018bayesian}.
Other examples of GANs for wind and PV power output scenarios are presented in \citep{jiang2018scenario,zhang2020typical,jiang2019forecasting}. 
For more consistent convergence in training, the authors in \cite{chen2018model, jiang2018scenario, zhang2020typical, jiang2019forecasting} used Wasserstein GANs \citep{arjovsky2017wasserstein} where they enforce a Lipschitz constraint on the critic network. 
Besides the generation of wind and PV scenarios, GAN-based scenario generation was also applied to residential load forecasts \citep{gu2019gan} and hydro-wind-solar hybrid systems \citep{wei2019short}. 
\cite{schreiber2019generative} study different loss functions for GANs and found the Wasserstein distance to be superior to the binary-cross-entropy. 
Besides GANs, a popular type of DGMs are variational autoencoders (VAEs) \citep{kingma2013auto}. 
Examples of VAEs for scenario generation include electric vehicle load demand \citep{pan2019data}, hydro wind-solar hybrid systems \citep{zhanga2018optimized}, as well as hydro concentrated solar power hybrid systems \citep{qi2020optimal}.

Despite their recent success in scenario generation for renewable power generation, both GANs and VAEs show inconsistencies in training.
Specifically, the Nash Equilibria obtained through GAN training are reported to be unstable, and there is no guarantee for the generator to sample from the target distribution \citep{arjovsky2017wasserstein}.
The VAE uses the Evidence Lower Bound (ELBO) loss function in training which gives no concrete measure on the actual quality of the fit \citep{kingma2013auto}.

In contrast to GANs and VAEs, a third DGM structure, normalizing flows \citep{dinh2016realNVP, papamakarios2019normalizing}, can directly fit the probability density function (PDF) of the unknown distribution by log-likelihood maximization.
Similar to the more established transport map approach \citep{Marzouk2016}, normalizing flows model the distribution of a data set as a deterministic transformation of a Gaussian. 
While transport maps typically use triangular maps \citep{Marzouk2016} to set up convex fitting problems, normalizing flows are designed using invertible neural networks to build more flexible designs. 
In theory, using a sufficiently expressive normalizing flow network and enough training data, the trained distribution will converge to the true distribution.
However, normalizing flows are not as well established as GANs and VAEs in scenario generation yet. To our knowledge, the only works using normalizing flows in the context of energy time series are \cite{zhang2019scenario} and \cite{ge2020modeling} focusing on demand time series and \cite{dumas2021deep} generating PV and wind electricity time series.

Recently, \cite{brehmer2020flows} and \cite{behrmann2021understanding} showed that normalizing flows are by design unable to fit distributions that lie on lower-dimensional manifolds. 
Instead, fitting a full dimensional normalizing flow to a manifold distribution results in exploding likelihood functions from numerically singular Jacobians \citep{behrmann2021understanding} and a smeared-out version of the true distribution \citep{brehmer2020flows}. Sampling from the smeared-out fit then leads to the samples outside of the true distribution, e.g., noisy time series.
In 2016, \cite{gemici2016normalizing} presented the necessary foundations to build normalizing flows on manifolds, i.e., to perform density estimation in a lower-dimensional space. However, they do not further elaborate on such an approach and do not report on any numerical experiments. 
\cite{brehmer2020flows} build normalizing flows in lower-dimensional space by fixing some of the latent space dimensions to a constant. 

We find manifolds to be frequently present in energy time series due to the temporal correlation between time steps. 
However, \cite{zhang2019scenario}, \cite{ge2020modeling}, and \cite{dumas2021deep} employ standard normalizing flow structures to such data, i.e., real non volume preserving transformation (RealNVP) \citep{dinh2016realNVP} or nonlinear independent component estimation (NICE) \citep{dinh2014nice}.
As a result, their works suffer from complications resulting from data manifolds. 
For instance, the training of \cite{ge2020modeling} results in extremely high density values ($\log p(x) \approx 4000$), which indicates that the model is trying to describe an infinite density. The validation loss shown by the authors indicates strong overfitting. However, the authors do not declare the exploding likelihood and the overfitting as issues and present the spurious results without further discussion. 
\cite{zhang2019scenario} acknowledge that their scenarios exhibit noisy behavior that does not match the expected results. However, they attribute this noise to their conditional training approach. 
\cite{dumas2021deep} draw a comparison between normalizing flows, GANs, and VAEs at the example of PV, wind, and load time series. The authors find that their normalizing flow models are unable to recover the autocorrelation within the time series and instead exhibit very high fluctuations. However, they still conclude that scenarios generated from normalizing flows show good results despite the unrealistic characteristics. 
Based on our observation of the frequent complications resulting from data manifolds, we argue that normalizing flows should be set up with a lower-dimensional latent space to avoid generating unrealistic energy time series scenarios. 

As part of this paper, we provide the following contributions:
We elicit the contradiction between manifolds and the diffeomorphic transformations required for normalizing flows. Further, we use simple two-dimensional examples to highlight how data manifolds lead to the generation of unrealistic and out-of-distribution data.
In addition, we show how the principal component analysis (PCA) can be used to reduce the dimensionality of energy time series effectively and show theoretically that the resulting principal component flow (PCF) does not affect the density estimation procedure, which avoids the need for balancing reconstruction and likelihood maximization losses like in \cite{brehmer2020flows}. 
In numerical experiments, we train the PCF on data of PV and wind power generation as well as load demand in Germany in the years 2013 to 2015. For reference, we compare our results with scenarios generated using the alternative methods Copulas \citep{pinson2009probabilistic} and Wasserstein GANs \citep{chen2018model}.
The results show that the PCF learns the distributions better than the full space normalizing flow (FSNF) and also maintains the frequency behavior of the original time series. The PCF-generated scenarios perform equally good or better compared to the Copula and W-GAN scenarios in all considered metrics. 

The remainder of this paper is organized as follows:
In Section \ref{sec:NormalizingFlows}, we introduce the general concept of normalizing flows and review the RealNVP affine coupling layer \citep{dinh2016realNVP} which will serve as the underlying flow structure in this paper.
In Section \ref{sec:PCA_Layer}, we continue the discussion started by \cite{brehmer2020flows} and \cite{behrmann2021understanding} about the effects of manifolds on normalizing flows through a clarifying toy example and show theoretically that PCA does not affect the density estimation in lower-dimensional latent space.
In Section \ref{sec:CaseStudies}, we present the results of simulation studies on data of PV and wind power generation as well as load demand in Germany \citep{DataSource}.
Finally, in Section \ref{sec:Conclusion} we conclude our work. 

%% file: sections/02_NormalizingFlows.tex
\section{Density Estimation using Normalizing Flows}\label{sec:NormalizingFlows}
Normalizing flows are invertible transformations $\mathbf{f}$ between a complex target distribution and a well-described base distribution, e.g., a multivariate Gaussian~\citep{papamakarios2019normalizing}. 
Analog to the inverse transport map methodology~\citep{Marzouk2016}, normalizing flows aim to model a complex distribution as a transformation of a simple one instead of manually deriving a complex model that fits the data. Thus, the sampling takes place in the known base distribution, and the transformation does not have to consider the randomness in the data. 
\begin{figure}
    \centering
    \includegraphics[width=0.8\textwidth]{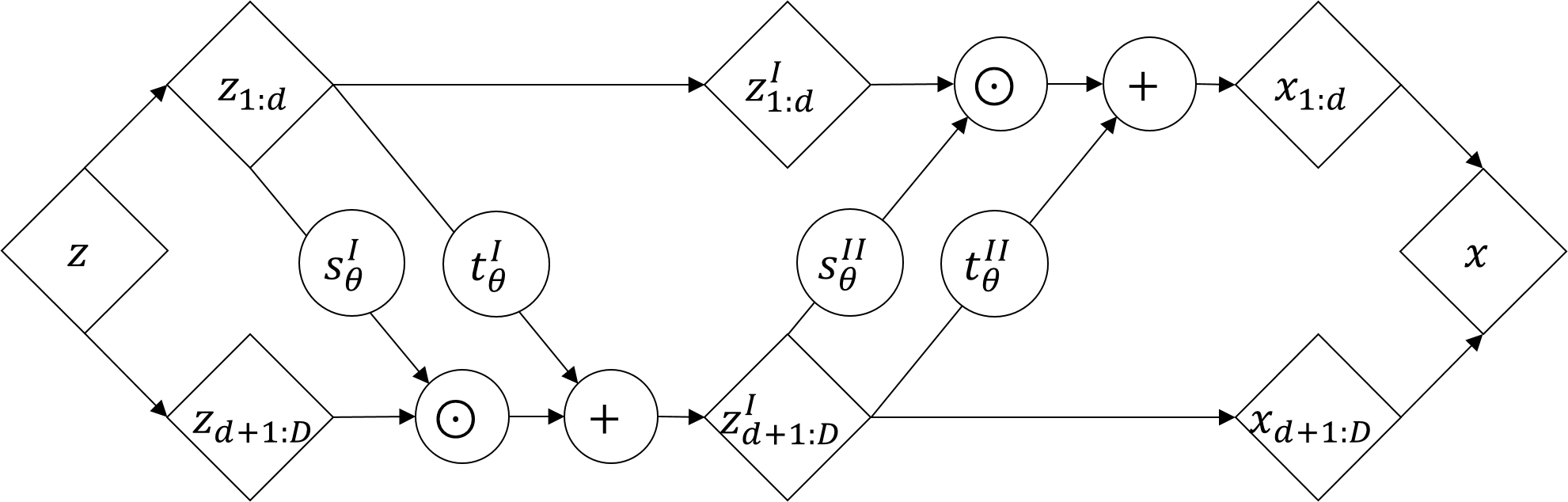}
    \caption{Real non-volume preserving transformation (RealNVP) \citep{dinh2016realNVP} with two coupling layers with alternating identity and affine transformations. Arrows point in generative direction. Compositions may include more than two coupling layers. Functions $\mathbf{s}^I_{\bm{\theta}}(\mathbf{z}_{1:d})$, $\mathbf{t}^I_{\bm{\theta}}(\mathbf{z}_{1:d})$, $\mathbf{s}^{II}_{\bm{\theta}}(\mathbf{z}^{I}_{d+1:D})$, and $\mathbf{t}^{II}_{\bm{\theta}}(\mathbf{z}^{I}_{d+1:D})$ are trainable ANNs with parameters $\bm{\theta}$.}
    \label{fig:realNVPsketch}
\end{figure}

A normalizing flow transformation must be set up as a diffeomorphism, i.e., both the forward and the inverse transformation must be continuously differentiable \citep{papamakarios2019normalizing}.
In their standard form, normalizing flows require equal dimensionality of base and target distributions, in which case the density of the target distribution is well described using the change of variables formula (CVF) \citep{papamakarios2019normalizing}
\begin{align}
    p_X (\mathbf{x}) = p_Z (\mathbf{f}^{-1} (\mathbf{x})) \left| \det \mathbf{J}_{\mathbf{f}}\left(\mathbf{f}^{-1}(\mathbf{x}) \right)\right|^{-1} ,
    \label{Eq:ChangeOfVariablesGeneral}
\end{align}
where $p_X(\mathbf{x})$ and $p_Z(\mathbf{z})$ are the densities of the samples $\mathbf{x}$ and $\mathbf{z}$ of the target distribution $X$ and the base distribution $Z$, respectively, and $\mathbf{J}_{\mathbf{f}^{-1}}(\mathbf{x})$ is the Jacobian of the inverse transformation $\mathbf{f}^{-1}$. 
To fit complex distributions, the diffeomorphic transformation $\mathbf{f}$ needs to be flexible and expressive. 
However, expressive and yet easily invertible functions with tractable Jacobian determinants are often difficult to engineer. 
Fortunately, diffeomorphisms are composable, and therefore normalizing flow models can be built using compositions of simple transformations, i.e.,
\begin{align}
    \mathbf{f} = \mathbf{f}_K \circ \mathbf{f}_{K-1} \circ \dots \circ \mathbf{f}_{2} \circ \mathbf{f}_{1},
\end{align}
where $\mathbf{f}_1$ to $\mathbf{f}_K$ are simple diffeomorphisms and the operator $\circ$ denotes function composition.
In logarithmic form, the CVF of compositions is given by
\begin{align}
        \log~p_X(\mathbf{x}) &= \log~p_Z (\mathbf{f}^{-1} (\mathbf{x})) + \sum_{k=1}^K \log \left| \det\mathbf{J}_{\mathbf{f}_k} \left( \mathbf{f}_k^{-1} (\mathbf{x}_{k}) \right)\right|^{-1}, \label{Eq:ChangeOfVariablesLog} 
\end{align}
where $\mathbf{x}_k$ is an intermediate variable $\mathbf{x}_k=\mathbf{f}_k(\mathbf{x}_{k-1})$.
In practice, the transformation is often set up as a trainable function $\mathbf{f}_{\bm{\theta}}$ with parameters ${\bm{\theta}}$.
By using the CVF, the transformation $\mathbf{f}_{\bm{\theta}}$ can be trained via direct likelihood maximization and for numerical reasons, the log form in Equation~\eqref{Eq:ChangeOfVariablesLog} is maximized in training:
\begin{align}
    \underset{{\bm{\theta}}}{\max}\quad\log~p_X(\mathbf{x};{\bm{\theta}}) = \log~p_Z (\mathbf{f}_{\bm{\theta}}^{-1} (\mathbf{x})) + \log  \left| \det \mathbf{J}_{\mathbf{f}_{\bm{\theta}}}\left(\mathbf{f}_{\bm{\theta}}^{-1}(\mathbf{x}) \right)\right|^{-1} 
\end{align}
Here, the likelihood $p_X(\mathbf{x};{\bm{\theta}})$ is parameterized by the trainable parameters ${\bm{\theta}}$ of the transformation $\mathbf{f}_{\bm{\theta}}$ and the historical samples of the target distribution $\mathbf{x}$ take the role of training data. 

In the last six years, many flow construction methods have been proposed, e.g., \cite{dinh2014nice} and \cite{dinh2016realNVP}. A prominent normalizing flow model is RealNVP \citep{dinh2016realNVP} which uses an affine coupling layer and has shown promising results in a prior application to time series data \citep{zhang2019scenario}. 
The idea of the RealNVP affine coupling layer is to split the full input vector $\mathbf{z}=\mathbf{z}_{1:D}$ of dimension $D$ and apply an affine transformation to one part of the input vector $\mathbf{z}_{d+1:D}$ conditioned on the remaining part of the input vector $\mathbf{z}_{1:d}$ that is kept constant.
Here, $d$ is usually set to $d=D/2$ to allow for maximal interaction between dimensions but can take other values $1<d<D$, e.g., if $D$ is uneven.  
The standard forward transformation $\mathbf{f}_{CL}: \mathbf{z}\rightarrow \mathbf{x}$ is given by
\begin{align}
    \mathbf{x}_{1:d} &= \mathbf{z}_{1:d}, \label{Eq:IdentityTransformation} \\
    \mathbf{x}_{d+1:D} &= \exp\left(\mathbf{s}_{\bm{\theta}}\left(\mathbf{z}_{1:d}\right)\right) \odot \mathbf{z}_{d+1:D} + \mathbf{t}_{\bm{\theta}}\left(\mathbf{z}_{1:d}\right), \label{Eq:AffineTransformation}
\end{align}
where the functions $\mathbf{s}_{\bm{\theta}}(\mathbf{z}_{1:d})$ and $\mathbf{t}_{\bm{\theta}}(\mathbf{z}_{1:d})$ are feed-forward ANNs called conditioner networks with parameters ${\bm{\theta}}$ and input and output dimensions $d$ and $D-d$, respectively. The $\odot$ operator denotes element-wise multiplication.
Note that when applied as a composition in alternating form RealNVP can build flexible and easily invertible transformations with tractable Jacobian determinants.
A visual description of a composition of two affine coupling layers is presented in Figure~\ref{fig:realNVPsketch}.

The affine coupling layer in Equations \eqref{Eq:IdentityTransformation} and \eqref{Eq:AffineTransformation} has the advantage that the Jacobian of the transformation $\mathbf{f}_{CL}$ is a lower triangular matrix, i.e., 
\begin{align}
    \mathbf{J}_{\mathbf{f}_{CL}}(\mathbf{z}) = \left[ \begin{array}{cc}
        \mathbf{I} & \mathbf{0}\\
        \frac{\partial \mathbf{x}_{d+1:D}}{\mathbf{z}_{1:d}} & \text{diag}[\exp(\mathbf{s}_{\bm{\theta}}(\mathbf{z}_{1:d}))] 
    \end{array}\right], \label{Eq:realNVPJacobian}
\end{align}
which means that the log of the absolute value of its determinant is simply given by
\begin{align}
    \log\left|  \det\left( \mathbf{J}_{\mathbf{f}_{CL}}(\mathbf{z})\right)  \right| = \sum^D_{i=d+1} \mathbf{s}_{{\bm{\theta}},i}(\mathbf{z}_{1:d}), \label{Eq:realNVPLogDetJacobian}
\end{align}
and the Jacobian of the inverse transformation $f_{CL}^{-1}: \mathbf{x}\rightarrow \mathbf{z}$ satisfies
\begin{align}
    \log\left|  \det\left( \mathbf{J}_{\mathbf{f}^{-1}_{CL}}(\mathbf{x})\right)  \right| = - \sum^D_{i=d+1} \mathbf{s}_{{\bm{\theta}},i}(\mathbf{x}_{1:d}), \label{Eq:realNVPLogDetJacobianInverse}
\end{align}
according to the inverse function theorem \citep{papamakarios2019normalizing}.
With Equations \eqref{Eq:realNVPLogDetJacobian} and \eqref{Eq:realNVPLogDetJacobianInverse}, the log-Jacobian determinant is computed from a simple evaluation of the forward or inverse transformation. 

For a more detailed introduction and a review of other normalizing flow designs, the interested reader is referred to the original RealNVP paper by \cite{dinh2016realNVP} and the review articles by \cite{papamakarios2019normalizing} and \cite{kobyzev2019normalizing}.

%% file: sections/03_PrincipalComponentFlow.tex
\section{Principal Component Density Estimation}\label{sec:PCA_Layer}
In this section, we first discuss the effects of manifolds on the Jacobians and the contradiction between manifolds and diffeomorphic transformations and visualize how distributions get smeared out at the example of a one-dimensional distribution embedded in two-dimensional space. 
Secondly, we show how unlike many other dimensionality reduction techniques PCA can be used in combination with standard normalizing flows for manifold density estimation with tractable and direct likelihood computation.

\subsection{Normalizing Flows and Manifolds}\label{sec:NFandMani}
Normalizing flows are transformations between the space of observable variables (target distribution) and a latent space with independent variables, i.e., a space with zero covariance (base distribution) \citep{papamakarios2019normalizing}. 
As a consequence, normalizing flows disentangle the information contained in the observable variables, e.g., the power generation over a given period, to the set of independent latent variables, i.e., the normalizing flow transformation eliminates the correlation between the dimensions \citep{papamakarios2019normalizing,kobyzev2019normalizing}. 
For the disentanglement to function properly, the normalizing flow transformation $\mathbf{f}$ and its inverse $\mathbf{f}^{-1}$ must be continuously differentiable, i.e., the Jacobian of $\mathbf{f}$ must be non-singular for $\mathbf{f}^{-1}$ to be differentiable and vise versa \citep{papamakarios2019normalizing}. 
The distributions of many data sets do not occupy the entire space of observable variables, i.e., the distributions lie on lower-dimensional manifolds \citep{fefferman2016testing}. 
A transformation from a latent space Gaussian with equal dimensionality to the space of observable variables must compress the Gaussian samples to the lower-dimensional space and, therefore, have a singular Jacobian \citep{hyvarinen1999nonlinear,behrmann2021understanding}. 
Consequently, a Gaussian with equal dimensionality cannot be matched to a manifold distribution using a diffeomorphic transformation.

However, normalizing flow transformations like RealNVP \citep{dinh2016realNVP} are designed to be diffeomorphic between spaces of equal dimensionality. When applied to manifold data, they have to include all latent space variables in the transformation with a non-zero gradient even if the data does not show any related variance. 
This leads to a contradiction between a problem that cannot be solved using a diffeomorphism and a strictly diffeomorphic normalizing flow. 
In practice, fitting a normalizing flow to manifolds leads to numerically singular Jacobians \citep{behrmann2021understanding} as the training attempts to compress the Gaussian to the lower-dimensional space. 
For sampling, the learned distribution is then smeared out around the true distribution \citep{brehmer2020flows}, causing skewed distribution densities and generation of out-of-distribution data. 

To illustrate this fundamental problem, we train two RealNVP models with identical architecture on samples of a one-dimensional, uniform distribution embedded in a two-dimensional curve and a two-dimensional distribution.
\begin{figure}
    \centering
    \includegraphics[width=\textwidth]{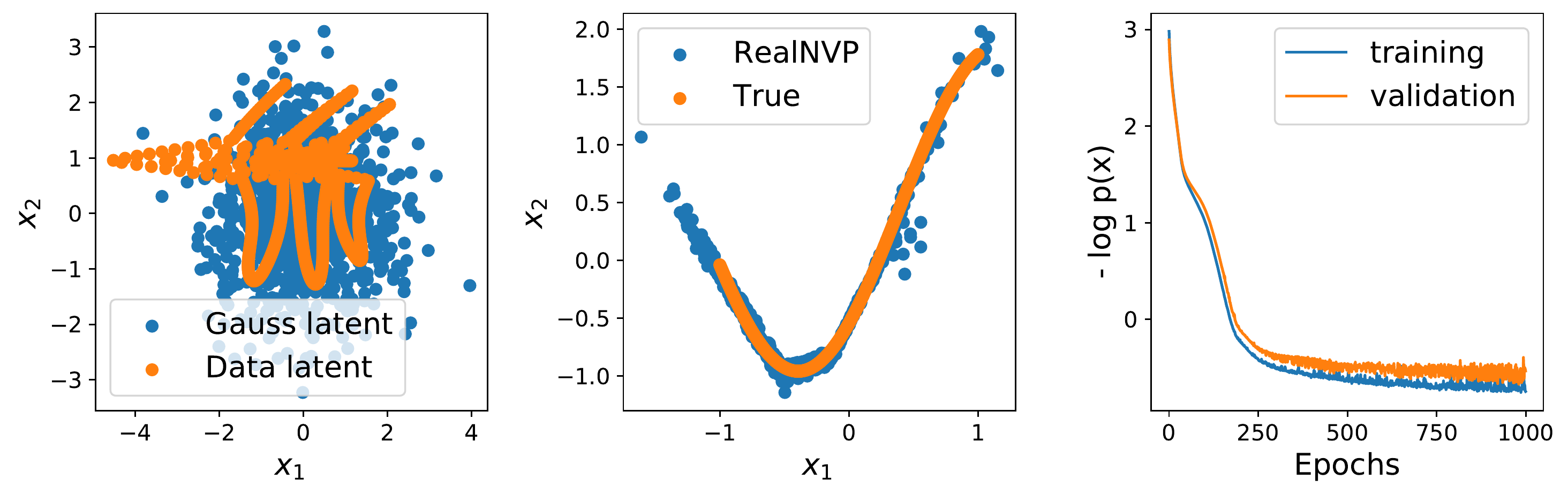}
    \caption{RealNVP~\citep{dinh2016realNVP} trained on 1D manifold in 2D spac e($x_1$, $x_2$). Left: samples of 2D Gaussian (blue) and training data after transformation to Gaussian (orange). Center: Samples from trained RealNVP (blue) and true (orange) data distribution. Right: training and validation loss over number of epochs.}
    \label{fig:MotivatingExample}
\end{figure}
\begin{figure}
    \centering
    \includegraphics[width=\textwidth]{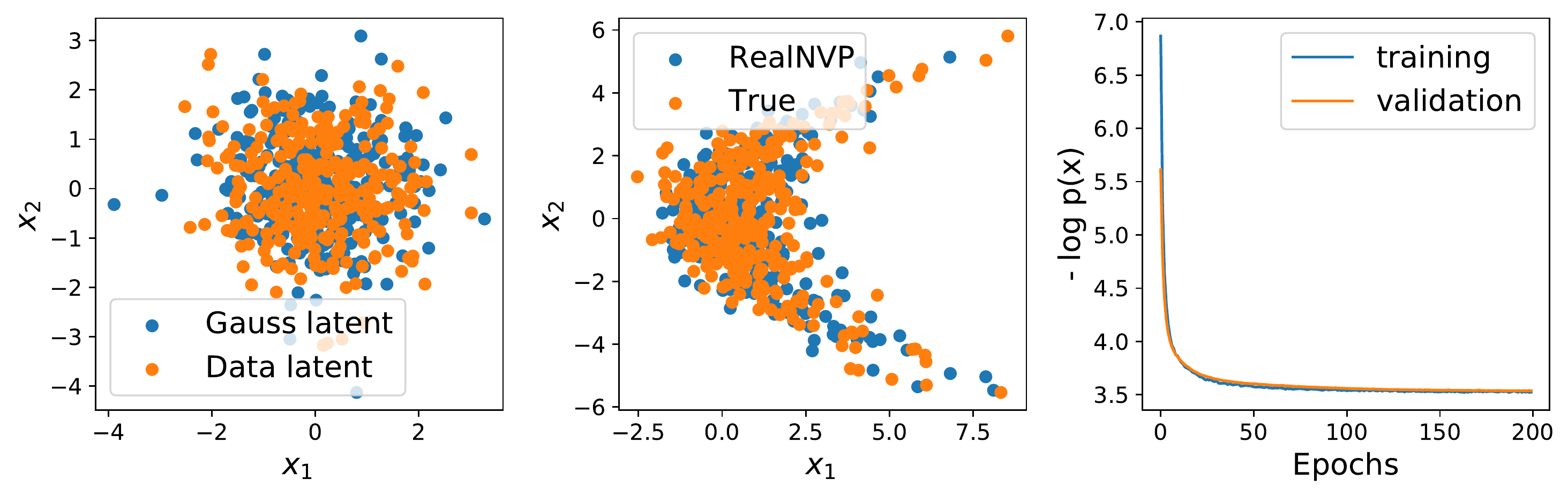}
    \caption{RealNVP~\citep{dinh2016realNVP} trained on 2D kite-shaped distribution in 2D space ($x_1$, $x_2$). Left: samples of 2D Gaussian (blue) and training data after transformation to Gaussian (orange). Center: Samples from trained RealNVP (blue) and true (orange) data distribution. Right: training and validation loss over number of epochs.}
    \label{fig:MotivatingExample2D}
\end{figure}
Figure~\ref{fig:MotivatingExample} and Figure~\ref{fig:MotivatingExample2D} show the encoded training data (Data latent) and samples of a two-dimensional Gaussian (Gauss latent) on the left, the training data (True), and samples generated using the RealNVP (RealNVP) in the center, and training and validation loss on the right for the two different distributions, respectively. 
The example in Figure~\ref{fig:MotivatingExample} highlights that the points from the one-dimensional curve remain on a one-dimensional curve after the transformation with the diffeomorphism. Accordingly, any point in the latent space outside of this curve has no counterpart in the true data distribution. 
Thus, the samples drawn in the Gaussian latent space (left) are transformed to out-of-distribution data, as can be seen by the RealNVP samples outside of the curve in the center of Figure~\ref{fig:MotivatingExample}. Nevertheless, the generated distributions resemble the shape of the true distributions and the out-of-distribution samples may not be noticed for higher-dimensional distributions, e.g., time series, that cannot be viewed in their ambient space. 

The absolute values of the loss functions are not comparable as they describe different distributions. However, while the validation loss of the one-dimensional example in Figure~\ref{fig:MotivatingExample} indicates overfitting and unstable training, the validation loss of the two-dimensional example in Figure~\ref{fig:MotivatingExample2D} indicates good generalization, stable training, and convergence in fewer epochs. 
Besides the loss functions, the data transformed to the latent space in the left of Figure~\ref{fig:MotivatingExample2D} and the RealNVP generated data in the center of Figure~\ref{fig:MotivatingExample2D} show good matches of the Gaussian and true data distribution, respectively.

\subsection{Principal Component Flow Layer}
Given an available injective map $\psi: \mathbb{R}^M \rightarrow \mathbb{R}^D, \mathbf{x}=\psi(\mathbf{\tilde{x}})$ with $D>M$ transforming data points from the lower-dimensional latent space $\mathbf{\tilde{x}}$ to their full space embedding $\mathbf{x}$, a flow on a Riemannian manifold can be built~\citep{gemici2016normalizing}.
Since the injective map expands the dimensionality, the typical CVF-relation (Equation~\eqref{Eq:ChangeOfVariablesLog}) of infinitesimal volumes using the Jacobian determinant does no longer apply.
Instead, the relation between the two infinitesimal volumes is given by
\begin{align}
    dX = \sqrt{\det(\mathbf{J}_\psi^T \mathbf{J}_\psi)}~dZ,
\end{align}
with Jacobian $\mathbf{J}_\mathbf{\psi}$ of the the mapping $\mathbf{\psi}$.
The generalized form of the CVF for the injective flow then is:
\begin{align}
    p_X(\mathbf{x}) = p_Z(\psi(\mathbf{x})^{-1}) \left[\left| \det\left(\mathbf{J}_{\psi}(\psi^{-1}(\mathbf{x}))^T \mathbf{J}_{\psi}(\psi^{-1}(\mathbf{x})) \right)\right|\right] ^{-0.5}\label{Eq:CVF_dimred}
\end{align}

For the injective map $\psi$ to be applicable for the use in combination with normalizing flows, its inverse $\psi^{-1}$ must be available and easy to compute, and the Jacobian determinant term in Equation~\eqref{Eq:CVF_dimred} must be computationally tractable during training. 
\cite{brehmer2020flows} propose to build an injective map by fixing some of the latent space dimensions of a normalizing flow to a constant. However, their approach still results in expensive Jacobian computations and requires balancing a reconstruction loss with the likelihood maximization.
To avoid expensive Jacobian computations and achieve easily invertible dimensionality reduction, we exploit the isometric, i.e., distance preserving, characteristic of the affine PCA \citep{pearson1901pca}.
PCA is based on the singular value decomposition of the sample covariance matrix $\mathbf{K}_{X,X}$ of the data distribution $X$, i.e.,
\begin{align}
    \mathbf{K}_{X,X}  = \frac{1}{N-1}\sum_{i=1}^N (\mathbf{x}_i - \bm{\mu}_X) (\mathbf{x}_i - \bm{\mu}_X)^T = \mathbf{U\Sigma V}^{-1},
\end{align}
where $N$ is the number of data points, $\mathbf{x}$ are the data points, and $\bm{\mu}_X$ is the empirical mean vector of the distribution. The covariance matrix $\mathbf{K}_{X,X}$ is decomposed into a diagonal matrix of singular values $\mathbf{\Sigma}$ and two unitary matrices $\mathbf{U}$ and $\mathbf{V}$ of left and right singular vectors, respectively.
The columns of the matrix of right singular vectors $\mathbf{V}$ corresponding to the largest singular values are called the principal components of the data distribution $X$. 
By truncating the columns with small or zero singular values, we can use the resulting semi-orthogonal matrix $\mathbf{V}_P$ for the affine embedding function:
\begin{equation}
    \mathbf{x} = \psi_{\text{PCA}}(\mathbf{\tilde{x}}) = \mathbf{V}_P\mathbf{\tilde{x}} + \bm{\mu}_X 
    \label{Eq:PCAtransform}
\end{equation}
with $\mathbf{V}_P$ as the truncated matrix of right singular vectors and $\bm{\mu}_X$ as the mean value of the distribution \citep{pearson1901pca}. 

In general, isometries are defined via:
\begin{equation}
    \mathbf{J}_{\psi}(\mathbf{\tilde{x}})^T \mathbf{J}_{\psi}(\mathbf{\tilde{x}}) = \mathbf{I}_M, 
    \quad \forall \mathbf{\tilde{x}} \in \mathbb{R}^M \label{Eq:IsomertyDefinition}
\end{equation}
Here, $ \mathbf{I}_M$ is the $M\times M $ identity matrix. 
For PCA, Equation~\eqref{Eq:IsomertyDefinition} holds as the Jacobian of $\psi_{\text{PCA}}$ is $\mathbf{J}_{\psi_{\text{PCA}}}(\mathbf{\tilde{x}}) = \mathbf{V}_P$, where $\mathbf{V}_P$ is semi-orthogonal. Thus:
\begin{equation}
    \mathbf{J}_{\psi_{\text{PCA}}}(\mathbf{\tilde{x}})^T \mathbf{J}_{\psi_{\text{PCA}}}(\mathbf{\tilde{x}}) = \mathbf{V}_P^T \mathbf{V}_P = \mathbf{I}_M, 
    \quad \forall \mathbf{\tilde{x}} \in \mathbb{R}^M \label{Eq:IsometryPCA}
\end{equation}

\begin{figure}
    \centering
    \includegraphics[width=0.7\textwidth]{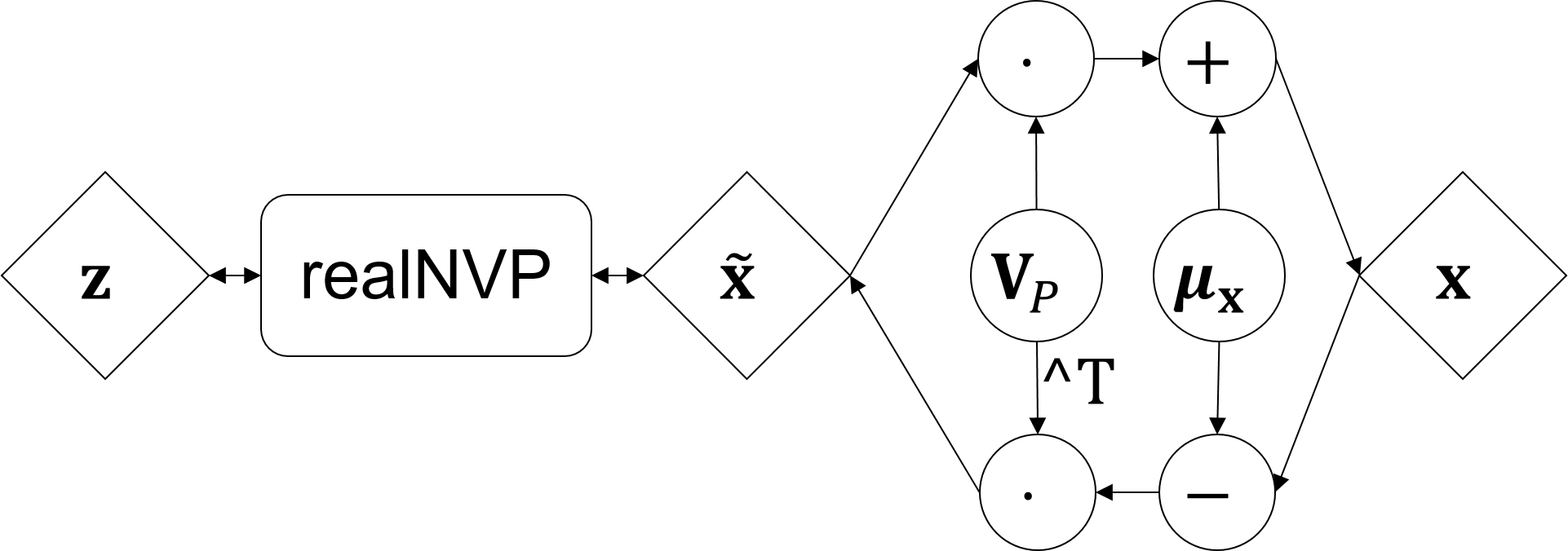}
    \caption{Principal component flow (PCF) structure with PCA layer as last layer in generative direction and RealNVP (see Figure \ref{fig:realNVPsketch}) as trainable normalizing flow in lower-dimensional space.}
    \label{fig:PCF}
\end{figure}

The isomertric property of PCA results in two major advantages for the combination with normalizing flows.
First, the PCA (pseudo-) inverse can be computed using the transpose of $\mathbf{V}_P$:
\begin{align}
    \mathbf{\tilde{x}} = \psi_{\text{PCA}}^{-1}(\mathbf{x}) = \mathbf{V}_P^T\left(\mathbf{x} - \bm{\mu}_X \right),
\end{align}
which makes the inverse explicit and computationally efficient.
Second, the PDF described by the CVF is invariant to the PCA dimensionality reduction, as the expression in Equation~\eqref{Eq:IsometryPCA} enters in the Jacobian determinant term of Equation~\eqref{Eq:CVF_dimred}, i.e.,
\begin{align}
    \left[\left| \det\left(\mathbf{J}_{\psi_{\text{PCA}}}(\mathbf{\tilde{x}})^T \mathbf{J}_{\psi_{\text{PCA}}}(\mathbf{\tilde{x}}) \right)\right|\right] ^{-0.5}
    = \left[\left| \det\left(
    \underbrace{\mathbf{V}_P^T \mathbf{V}_P}_{=\mathbf{I}_M}
    \right)\right|\right]^{-0.5} = 1, \quad \forall \mathbf{\tilde{x}} \in \mathbb{R}^M.
    \label{Eq:PCAlogdet}
\end{align}
When we build a composition of a standard normalizing flow and the PCA following Equation~\eqref{Eq:ChangeOfVariablesLog} it becomes apparent that the PCA transformation does not influence the PDF described by the CVF:
\begin{equation}
\begin{aligned}
    \log~p_X(\mathbf{x}) 
    & = \log~p_Z \left(\mathbf{f}^{-1} \circ \psi_{\text{PCA}}^{-1}(\mathbf{x})\right) 
     +\log \left| {\det}\left(\mathbf{J}_{\mathbf{f}^{-1}}\left(\psi_{\text{PCA}}^{-1}(\mathbf{x})\right) \right)\right|
     + \underbrace{\log\left[ \left|\det\left(\mathbf{V}_P^T \mathbf{V}_P \right)\right|\right]^{-0.5} }_{=0} \\
    & = \log~p_Z \left(\mathbf{f}^{-1} (\mathbf{\tilde{x}})\right) +\log \left| {\det}\left(\mathbf{J}_{\mathbf{f}^{-1}}(\mathbf{\tilde{x}}) \right)\right| \label{Eq:CVF_PCA}
\end{aligned}
\end{equation}
In Equation~\eqref{Eq:CVF_PCA}, $\mathbf{\tilde{x}}=\psi_{\text{PCA}}^{-1}(\mathbf{x})$ is the lower dimensional representation (Equation~\eqref{Eq:PCAtransform}) and $\mathbf{f}$ is a standard normalizing flow model, e.g., RealNVP \citep{dinh2016realNVP}.
Because PCA does not influence the CVF, it can be solved prior to fitting the normalizing flow and the PCF layer does not include trainable variables for the log-likelihood maximization. Hence, we avoid composite loss functions that balance reconstruction loss and likelihood maximization as in \cite{brehmer2020flows}.
The available inverse and separation of PCA fit and normalizing flow training make the use of PCA computationally very efficient, as neither the inverse nor the Jacobian have to be computed at any point. Note that the efficient inverse and the omittable Jacobian determinant do not transfer to most other dimensionality reduction techniques, e.g., local linear embeddings \citep{roweis2000nonlinear} or diffusion maps \citep{coifman2006diffusion}. 
A graphical description of the combination of the PCA and RealNVP is shown in Figure \ref{fig:PCF}.

%% file: sections/04_NumericalExperiments.tex
\section{Numerical Experiments}\label{sec:CaseStudies}
In this section, we train the PCF on real-world time series data. The three different data sets considered contain data of PV power generation, wind power generation, and load demand, where each represents the total values in Germany in the years 2013 to 2015 \citep{DataSource}. 
In the analysis below, we treat each of the time series independently. 
Prior to any processing, we clean the data of any days with missing values. To our knowledge, there are no curtailment effects in the data. 
PV and wind scenarios are scaled by the installed capacity at each time step. Thus, the networks are trained on the so-called capacity factor. The load demand data is scaled to the $[0,1]$ interval resulting in what we refer to as the demand factor in the following. 
Learning the capacity factor distribution avoids skewness of the distribution due to the addition of further generation capacity in the given time frame. We have not observed long-term trends in the load demand data of the selected period. 
To be processed by the normalizing flow, the univariate time series are cut into intervals of one day each. With sampling intervals of 15\,min, this leads to a set of 96-dimensional time series fragments that will serve as training data, i.e., a set of 96-dimensional vectors, which is a typical approach in the field of DGM scenario generation \citep{chen2018model, zhang2019scenario}.
This approach is known as multivariate prediction \citep{ziel2018day} as multiple time steps are predicted at the same time. 

For a comparison with more established scenario generation approaches, we also generate scenarios using Copulas and Wasserstein GANs (W-GANs). In contrast to autoregressive models \citep{sharma2013wind} and similar to normalizing flows, Copulas, and W-GANs describe the full distribution independent of conditional inputs.
We use the linear interpolation of the inverse cumulative distribution function to fit the Copula to the data \citep{pinson2009probabilistic}, and for the W-GANs, we follow the approach by \cite{chen2018model}.
Note that both Copulas and W-GANs use the same idea of multivariate scenario generation, i.e., treating scenarios as a vector rather than a sequence of values. 

\begin{table}
    \centering
    \caption{Number of principal components for cumulative explained variance (CEV). Data with 15\,min resolution (96 dimensions) from \cite{DataSource}.}
    \begin{tabular}{rrrr}
        \hline
        Explained variance & PV     & wind & load demand \\ \hline
        $\geq$ 99.00\%     & 3      & 6    & 5 \\
        $\geq$ 99.90\%     & 6      & 10   & 16 \\
        $\geq$ 99.99\%     & 16     & 44   & 63 \\
        100.00\%           & 62     & 96   & 96 \\ \hline
    \end{tabular}
    \label{tab:ExplainedVariance}
\end{table}
First, we investigate the manifold dimensionality of the data sets by looking at the explained variance ratio of the principal components. 
The explained variance ratio is the value of the singular values of the covariance matrix scaled by their sum, i.e., the relative amount of variance in the data, which is described by the right singular vector (the principal component) corresponding to the given singular value \citep{tipping1999probabilistic}.
As an indicator of the manifold dimensionality, we look at the cumulative explained variance (CEV) which describes how much of the information in terms of variance is maintained when the data is compressed to the space of principal components.
Table~\ref{tab:ExplainedVariance} lists some relevant CEV values and the corresponding latent space dimensionality for our data sets. 
The results show that a significantly reduced dimensionality is sufficient to maintain close to all of the variance information in the data.  
The 100\,\% threshold is reached when the latent dimensionality is equal to the numerical rank of the covariance matrix. The PV data reaches this threshold at a latent dimensionality of $62$. 
Note that the 100\% CEV appears with fewer principal components than dimensions of the data, if either one or more of the dimensions has zero variance or if there is an exact linear dependency of one of the dimensions on one or more of the other dimensions. 
Another important thing to consider is that the PCA is fitted using the \textit{empirical} covariance matrix, i.e., with a finite number of samples. Hence, the numerical rank might be different from the rank of the covariance matrix of the process and the low variance components identified by the PCA may not be fully representative of the underlying process. Hence, the low variance and high-frequency components identified by the PCA may not be desired.

We train full-space normalizing flow (FSNF) and PCF models on each of the three different data sets and compare the generated to the historical data (target). 
For each data set, we select two latent dimensionalities, namely, 16 and 62 for PV to represent 99.99\% and 100\% of the variance, and 6 and 10 for wind as well as 5 and 16 for demand to represent 99\% and 99.9\% of the variance, respectively.
In each training, 20\% of the 1096 scenarios are set aside as validation sets for hyperparameter tuning and to avoid overfitting. All of the following analysis is performed using the full historical scenario sets in comparison to sets of generated scenarios, to avoid random errors in the evaluation as a result of too few data points.

We find that, for all networks, five RealNVP affine coupling layers with fully connected conditioner networks with two hidden layers of the same size as the input dimension each is sufficient and additional complexity does not improve the representation. All normalizing flow models are implemented using the Python-based machine learning libraries Tensorflow version 2.4.0 \citep{tensorflow2015} and Tensorflow-Probability version 0.12.1 \citep{dillon2017tensorflow}. We use the PCA routine from the open-source scikit-learn Python library version 0.24.0 \citep{scikitlearn} for dimensionality reduction.

\begin{figure}
    \centering
    \includegraphics[width=\textwidth]{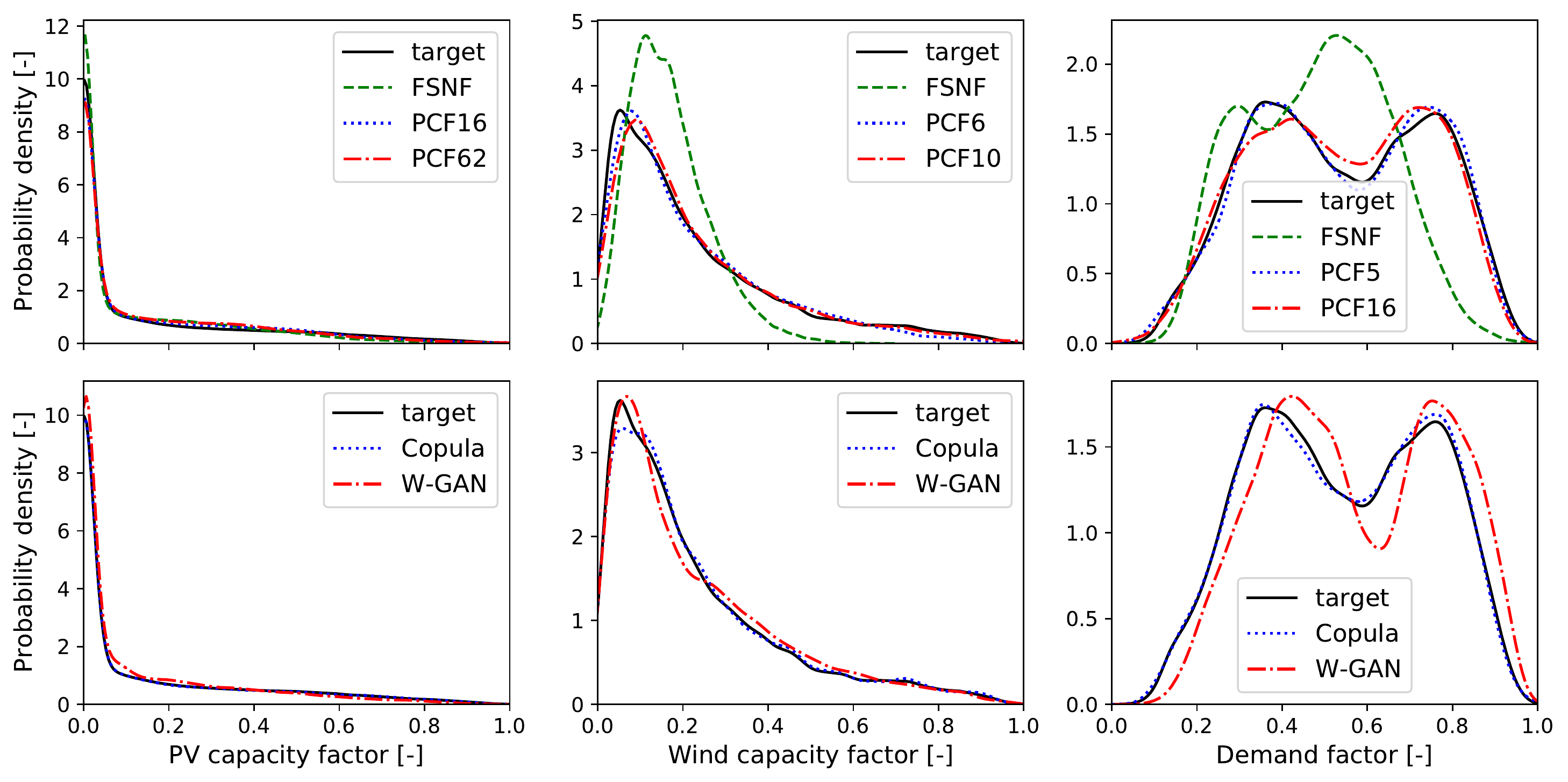}
    \caption{Comparison of the probability density function (PDF) from kernel density estimation \citep{parzen1962estimation}. 
    Top: historical data (target, solid lines), generated data from full space normalizing flow (FSNF, dashed lines), and generated data from PCF (PCF, dotted line and dash-dotted lines) 
    Bottom: historical data (target, solid lines), generated data from the Copula (Copula, dotted lines), and generated data from W-GAN (W-GAN, dash-dotted lines). 
    Left: PV capacity factor, 
    center: wind capacity factor, 
    right: demand factor.}
    \label{fig:PDF}
\end{figure}
We estimate the probability density function (PDF) of the data using kernel density estimation (KDE) with Gaussian kernels \citep{parzen1962estimation} as shown in Figure~\ref{fig:PDF}. The top line of Figure~\ref{fig:PDF} shows the PDF plots of the historical scenarios, FSNF-generated scenarios, and scenarios generated using the proposed PCF for each of the three data sets. For reference, Figure~\ref{fig:PDF} compares the Copula and W-GAN-generated scenario PDFs with the historical scenarios, in the second line.

For the PV data in the left of Figure~\ref{fig:PDF}, the data from all three normalizing flow structures appear to match the distribution. 
The FSNF data exhibits a higher density than the target for capacity factors smaller than $0.05$, between $0.1$ and $0.45$, and a lower density above a certain capacity factor of $0.45$. 
The PCF62 data shows an overestimation between a capacity factor of $0.1$ and $0.4$, however, the PCF62 density is closer to the density of the target distribution than the FSNF density, in particular, for capacity factors below $0.1$. 
Finally, the PCF16 data distribution is closest to the target distribution with good fits for capacity factors smaller than $0.2$ and similar deviations from the target as observed with the FSNF and the PCF62. 

For the wind capacity factor (center in Figure~\ref{fig:PDF}) and the demand factor (right in Figure~\ref{fig:PDF}), we observe a more drastic overestimation of the PDF for the highest densities and an underestimation of the tails by the FSNF. 
As for the PV scenarios, the density estimates for wind and load demand scenarios show an improved approximation of the distribution through dimensionality reduction with PCA, although the covariance matrices are not rank deficient.
It appears that the FSNF tends to ignore rare events during the likelihood maximization and thus overestimates the areas of high density.
With a lower dimensionality, rare events occupy a relatively larger space and are therefore more likely to be considered by the PCF likelihood maximization. 
Furthermore, the PDF fits of both wind and demand appear to improve by considering fewer principal components. 
By adding more principal components to get from $99\%$ to $99.9\%$ CEV, the training problem becomes less well-conditioned as the additional dimensions are only narrowly occupied by data, i.e., the distribution comes closer to a lower-dimensional manifold. 

The bottom line of Figure~\ref{fig:PDF} shows good matches of PV and wind capacity factor PDFs for both Copulas and W-GANs. The Copula scenarios also show a matching demand factor PDF, while the W-GAN results in a skewed PDF. 
A comparison of the top and bottom lines in Figure~\ref{fig:PDF} highlights that PDF representations by the PCF are as good or better compared to the literature approaches. 

Note that an exact evaluation of the KDE plot can be misleading since the KDE results are not exact, in particular, at the tails of the distribution and in low-density parts \citep{wied2012consistency}. 
However, the trends for all three data sets show improved fits of the PCF setups compared to the FSNF, which indicates better distribution fits due to the dimensionality reduction.

To support the results from the visual comparison of KDE plots, we run a Kolmogorov–Smirnov test (KS-test) \citep{Hodges1958TheSP}. The KS-test is used to judge whether two distinct sets of samples stem from the same distribution. 
Table~\ref{tab:KS_Test} lists the p-values (p-value $\in [0,1]$) of the tests. In general, high p-values ($\geq 0.1$) can be interpreted as a match of distributions, and low p-values indicate that the two sample sets stem from different distributions \citep{Hodges1958TheSP}.
\input{sections/88_KS_Table}
For all three data sets, the p-values of the FSNF indicate that the historical and the generated data do not follow the same distribution. 
In contrast, the p-values for PCF16 for PV, PCF6 for wind, and PCF5 for load demand are $\geq 0.1$, indicating a good representation of the historical data distribution. The PCF10 for wind and the PCF16 for demand show p-values $\geq 0.05$ and $\geq 0.01$ which still indicate good matches of the distribution and significantly better matches compared to the FSNF. 
The PCF62 for PV gives a lower p-value and hence lower confidence that the data stems from the historical data distribution. This result matches the observations in Figure~\ref{fig:PDF}, where we observed a slightly worse fit from PCF62 compared to PCF16. Still, there is a significant improvement compared to the FSNF. Furthermore, the KS-test confirms our observation of increasing quality of fit for the $99\%$ CEV compared to the $99.9\%$ CEV. 

The KS-test of the reference models, returns high p-values for all datasets for the Copula-generated scenarios, while only the wind capacity factor shows a high p-value among the W-GAN-generated scenarios. In general, the KS-test indicates good matching distributions only for Copulas and for the PCFs with low numbers of principal components.

In conclusion, the PCF returns significantly better fits of the distribution compared to the FSNF. Furthermore, the PDF fits by the PCF are at least as good or better compared to the literature methods. Considering fewer principal components appears to enhance the learning of the distributions, and accepting CEV values around 99\% does not compromise the overall match of the distribution.

\begin{figure}
    \centering
    \includegraphics[width=\textwidth]{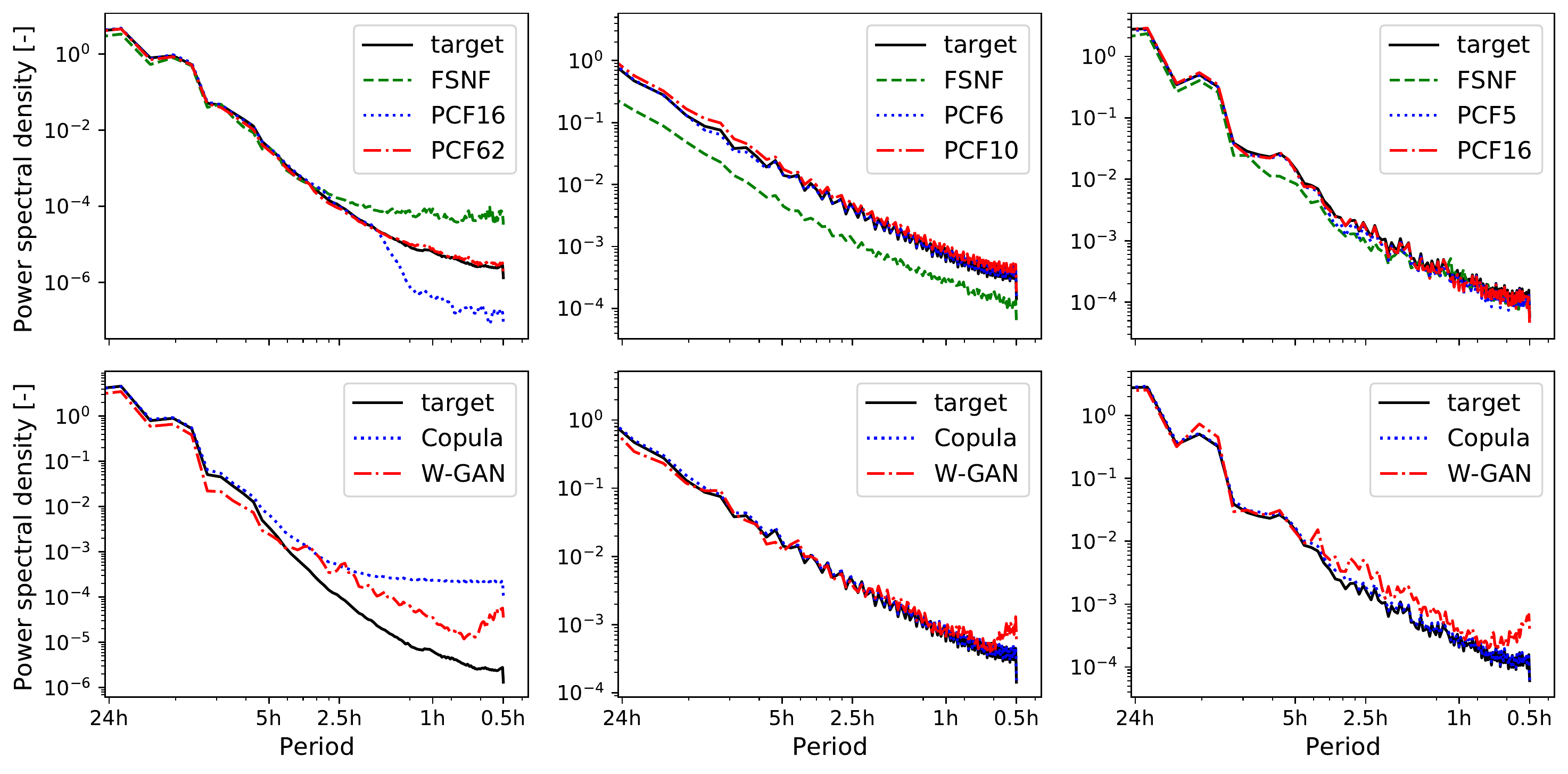}
    \caption{ 
    Power spectral density with Welch transform \citep{welch1967use}.
    Top: historical data (target, solid lines), generated data from full space normalizing flow (FSNF, dashed lines), and generated data from PCF (PCF, dotted line and dash-dotted lines). 
    Bottom: historical data (target, solid lines), generated data from the Copula (Copula, dotted lines), and generated data from W-GAN (W-GAN, dash-dotted lines). 
    Left: PV capacity factor, 
    center: wind capacity factor, 
    right: demand factor.
    }
    \label{fig:PSD}
\end{figure}

We use the power spectral density (PSD) based on the Welch transform \citep{welch1967use} to investigate in the frequency domain whether the FSNF and the PCF can reproduce the fluctuational behavior of the original time series. 
Figures~\ref{fig:PSD} shows the PSD of the target distribution, the FSNF, and the PCF generated scenarios for the three considered data sets, respectively. 

For the PV data, the target distribution shows high amplitudes ($>10^{-4}$) for low frequencies up to $\frac{1}{3.5\,h}$ which are well-matched by all three DGMs.
For higher frequencies ($>\frac{1}{3.5\,h}$), the target amplitude declines faster than the FSNF amplitudes, indicating the generation of noisy behavior over shorter periods by the FSNF. 
In contrast, PCF16 produces lower amplitudes than the target distribution in the regime of frequencies $>\frac{1}{2\,h}$, which indicates the filtering of some of the low amplitude fluctuation in the historical data set. 
The PCF62 shows a good fit for the true behavior of the target up to the highest frequencies. 

The PSD analysis for the wind data shows a good match of the frequency behavior for the data generated from the PCF6 and PCF10. The FSNF generated scenarios show a general underestimation of the amplitudes.
Note that the wind data exhibits higher fluctuation on all frequencies compared to the PV data. Therefore, the fluctuations at high-frequencies are represented in the principal components, and no filter effect can be observed. 
As the overall PSD shape is matched, the FSNF appears to find the right fluctuation behavior only with consistently lower amplitudes. 
Here, the narrower distribution described by the FSNF (see Figure~\ref{fig:PDF}) leads to a smaller range of scenarios and, therefore, lower amplitudes. 

As for the wind scenarios, the PSD analysis of the load demand data in Figure~\ref{fig:PSD} shows a good match of the PCF-generated scenarios to the historical data. The FSNF matches the overall PSD, but there are lower amplitudes between frequencies of $>\frac{1}{8\,h}$ and $>\frac{1}{2\,h}$.

The PSDs of the W-GAN-generated scenarios show an increase in amplitude for frequencies between $\frac{1}{1\,h}$ and $\frac{1}{0.5\,h}$ for all three datasets, indicating noisy scenarios. Furthermore, the PSDs of the PV capacity factor and the demand factor of the W-GAN-generated scenarios are higher compared to the historical scenario PSDs, i.e., the W-GANs generate scenarios with unrealistically high fluctuations.
The Copula-generated scenarios reproduce the PSDs of the wind capacity factor and the demand factor well. However, the Copula-generated PV capacity factor scenarios show high amplitudes for frequencies $>\frac{1}{5\,h}$. 

In conclusion, the PSD analysis of the PCF-generated scenarios clearly shows improved results for all data sets due to the dimensionality reduction compared to the FSNF.
Some effects are specific to the PCA dimensionality reduction of the given data, e.g., the filter effect observed for the PV data and the noise in the case of the load demand data. 
Among all considered scenario generation approaches, the PCF is the only approach that does not result in noisy behavior or otherwise higher fluctuations for all three datasets.

\begin{figure}
    \centering
    \includegraphics[width=\textwidth]{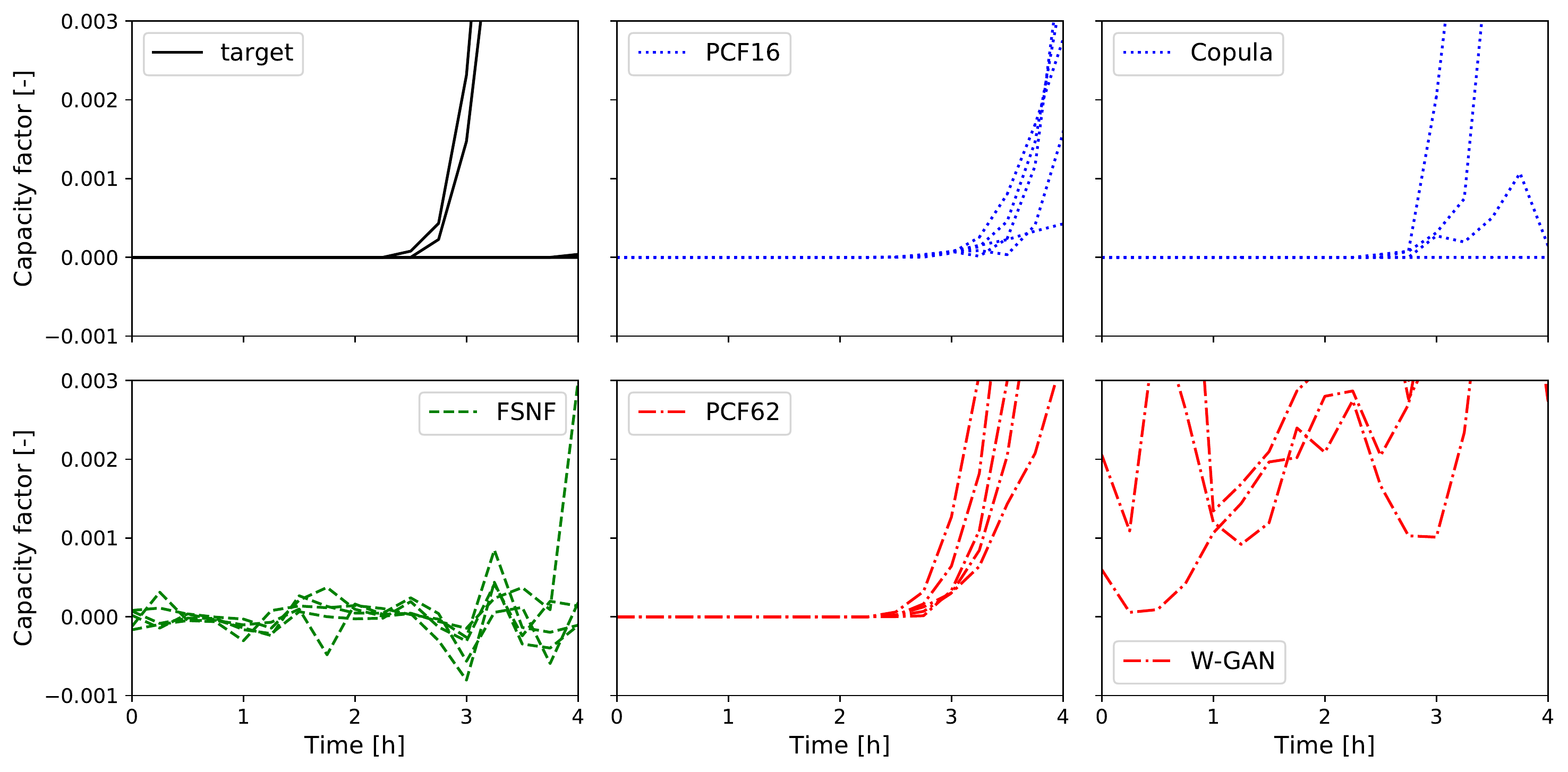}
    \caption{Five PV capacity factor scenarios from target, FSNF, PCF16, PCF62, Copula, and W-GAN respectively. Time frame between midnight and 4\,am.}
    \label{fig:nighttime}
\end{figure}
Finally, we investigate whether the PCF recovers the dimensions with zero variance that occur during the nighttime hours of the PV scenarios. 
Figure~\ref{fig:nighttime} shows the time between midnight and 4\,am of five randomly selected scenarios from the historical data, the normalizing flow models, and Copula and W-GAN-generated scenarios. 
The FSNF-generated scenarios clearly show noisy behavior. On the other hand, PCF16 and PCF62 do not show any noise and thus preserve the zero variance feature of the respective dimensions. 
The W-GAN scenarios show high fluctuations without any structure, while the Copula scenarios appear to identify the zero variance of the nighttime hours correctly. 
\input{sections/89_MeanVarTables}
In addition to Figure~\ref{fig:nighttime}, Table~\ref{tab:MeanVarSolar} lists the mean and variance for the time steps between midnight and 4\,am. The mean and variance values for PCF16 and PCF62 confirm the observations from Figure~\ref{fig:nighttime}. 
Figure~\ref{fig:nighttime} and Table~\ref{tab:MeanVarSolar} highlight the 'smeared-out' characteristic of the learned distribution as described in Section~\ref{sec:NFandMani} and by \cite{brehmer2020flows}. 
The results show that the FSNF is unable to detect dimensions of zero variance and does not accurately fit the manifold distribution, while PCF detects the manifold and correctly reproduces its features. 
In contrast to Figure~\ref{fig:nighttime}, the mean and variance values of the Copula-generated scenarios highlight that the Copula does not identify the zero variance characteristic of the nighttime hours. It appears that the Copula reproduces the zero variance characteristic for most but not all of the scenarios. 

In conclusion, the PCF is the only method among the considered approaches that consistently identifies the zero variance characteristic of the PV scenarios during nighttime. 

%% file: sections/88_KS_Table.tex
\begin{table}
    \centering
    \caption{
    P-values (p-value $\in [0,1]$) of Kolmogorov–Smirnov test \citep{Hodges1958TheSP}: 
    Statistical comparison of historical and generated data. High p-values ($\geq 0.1$) indicate high probability that generated scenarios have same distribution as the historical scenarios.  Results indicating a good match of the distributions are marked with an asterisk $^{*}$. 
    }
    \label{tab:KS_Test}

    \resizebox{\textwidth}{!}{%
\begin{tabular}{lllllllll}
\hline
       & FSNF                  & W-GAN                 & Copula       & PCF5         & PCF6         & PCF10  & PCF16        & PCF62                \\ \hline
Solar  & 2.67$\times 10^{-22}$ & 1.21$\times 10^{-04}$ & 0.4911$^{*}$ &              &              &        & 0.2483$^{*}$ & 2.07$\times 10^{-5}$ \\
Wind   & 6.69$\times 10^{-20}$ & 0.268$^{*}$           & 0.4782$^{*}$ & 0.9502$^{*}$ &              & 0.0537 &              &                      \\
Demand & 3.85$\times 10^{-81}$ & 6.35$\times 10^{-21}$ & 0.7727$^{*}$ &              & 0.2380$^{*}$ &        & 0.0361       &           \\ \hline          
\end{tabular}%
}
\end{table}

%% file: sections/89_MeanVarTables.tex
\begin{table}
\centering
\caption{Marginal mean and variance values for PV scenarios between midnight and 4\,am. Historical scenarios in comparison to FSNF, PCF16, and PCF62-generated scenarios as well as Copula and W-GAN generated scenarios.}
\label{tab:MeanVarSolar}

\resizebox{\textwidth}{!}{%
\begin{tabular}{l|llllll|llllll}
\hline
     & \multicolumn{6}{c}{Mean}                                                & \multicolumn{6}{c}{Variance}                                                 \\ \hline
     & Historical & FSNF                    & PCF16     & PCF62     & Copula    & W-GAN     & Historical & FSNF        & PCF16      & PCF62      & Copula     & W-GAN      \\ \hline
0:00 & 0          & -3.94$\times 10^{-5}$   & 0         & 0         & 9.88$\times 10^{-5}$  & 0.0046949 & 0          & -3.94$\times 10^{-5}$   & 0          & 0          & 9.88$\times 10^{-5}$   & 0.00469491 \\
0:15 & 0          & 4.91$\times 10^{-5}$    & 0         & 0         & 0.0018714 & 0.0030314 & 0          & 4.91$\times 10^{-5}$    & 0          & 0          & 0.00187143 & 0.00303145 \\
0:30 & 0          & 3.89$\times 10^{-5}$    & 0         & 0         & 0.000323  & 0.0048844 & 0          & 3.89$\times 10^{-5}$    & 0          & 0          & 0.00032296 & 0.00488435 \\
0:45 & 0          & -7.77$\times 10^{-5}$   & 0         & 0         & 0.0007804 & 0.0029591 & 0          & -7.77$\times 10^{-5}$   & 0          & 0          & 0.00078039 & 0.0029591  \\
1:00 & 0          & -0.0001856              & 0         & 0         & 0.0005853 & 0.0030573 & 0          & -0.00018558 & 0          & 0          & 0.00058527 & 0.00305727 \\
1:15 & 0          & -2.28$\times 10^{-5}$   & 0         & 0         & 0         & 0.0028363 & 0          & -2.28$\times 10^{-5}$   & 0          & 0          & -1.00$\times 10^{-10}$  & 0.00283627 \\
1:30 & 0          & 0.0001475               & 0         & 0         & 0         & 0.0028889 & 0          & 0.00014755  & 0          & 0          & -8.00$\times 10^{-10}$  & 0.00288887 \\
1:45 & 0          & 0.0001189               & 0         & 0         & 0.0005246 & 0.0032642 & 0          & 0.0001189   & 0          & 0          & 0.00052463 & 0.00326423 \\
2:00 & 0          & -1.14$\times 10^{-5}$   & 0         & 0         & 0         & 0.0030402 & 0          & -1.14$\times 10^{-5}$   & 0          & 0          & 6.00$\times 10^{-10}$   & 0.00304017 \\
2:15 & 0          & -1.81$\times 10^{-5}$  & 0         & 0         & 0         & 0.0031055 & 0          & -1.81$\times 10^{-5}$   & 0          & 0          & 6.00$\times 10^{-10}$   & 0.00310545 \\
2:30 & 4.00$\times 10^{-6}$ & 9.91$\times 10^{-5}$   & 3.10$\times 10^{-6}$  & 1.64$\times 10^{-5}$  & 0.0008303 & 0.0028191 & 4.04$\times 10^{-6}$   & 9.91$\times 10^{-5}$    & 3.05$\times 10^{-6}$   & 1.64$\times 10^{-5}$   & 0.00083027 & 0.00281913 \\
2:45 & 4.21$\times 10^{-5}$   & -0.0001062 & 3.31$\times 10^{-5}$  & 0.0001311 & 0.0005581 & 0.0029713 & 4.21$\times 10^{-5}$   & -0.00010616 & 3.31$\times 10^{-5}$   & 0.00013112 & 0.00055814 & 0.00297128 \\
3:00 & 0.0002261  & -0.0002776 & 0.0001789 & 0.0004552 & 0.0002479 & 0.0029983 & 0.00022609 & -0.00027757 & 0.00017895 & 0.00045525 & 0.00024788 & 0.00299829 \\
3:15 & 0.0006365  & 0.000326   & 0.0005092 & 0.0010372 & 0.0013232 & 0.0037313 & 0.00063655 & 0.00032604  & 0.00050917 & 0.00103717 & 0.00132324 & 0.00373134 \\
3:30 & 0.0016171  & 0.000278   & 0.0013039 & 0.0022794 & 0.0026894 & 0.0047879 & 0.00161713 & 0.00027802  & 0.00130388 & 0.00227943 & 0.0026894  & 0.00478792 \\
3:45 & 0.0032558  & 0.0006409  & 0.0026619 & 0.0040074 & 0.0040934 & 0.0062684 & 0.00325582 & 0.00064086  & 0.00266187 & 0.00400742 & 0.00409343 & 0.00626838 \\
4:00 & 0.0058694  & 0.0021747  & 0.0049369 & 0.0066868 & 0.0068678 & 0.009123  & 0.0058694  & 0.00217466  & 0.00493695 & 0.00668681 & 0.0068678  & 0.00912296 \\ \hline
\end{tabular}%
}
\end{table}

%% file: sections/05_Conclusion.tex
\section{Conclusion}\label{sec:Conclusion}
In this paper, we illustrate the issues associated with learning the probability distribution of electricity time series scenarios using normalizing flow density models. 
We find that these distributions lie on lower-dimensional manifolds due to the high autocorrelation between time steps. 
We show that standard normalizing flows are ill-suited to learn these manifold distributions and, instead, induce ill-conditioned or singular Jacobians that interfere with the normalizing flow training. However, this problem was previously not addressed in the literature on scenario generation, and we argue that this has led to some spurious results.
To mitigate the problems arising from data manifolds, we use dimensionality reduction.
Here, we exploit the isometric characteristic of the principal component analysis (PCA).
We show theoretically that the PCA does not influence the density estimation since its log-Jacobian determinant is by design always zero. Thus, we avoid having to balance reconstruction loss and likelihood maximization. Furthermore, no expensive Jacobian or matrix inverse computations are necessary, which makes PCA very computationally efficient. Critically, neither the explicit inverse nor the omittable Jacobian determinant apply to most other dimensionality reduction techniques. 

We apply the combined principal component flow (PCF) next to a standard full space normalizing flow (FSNF) to learn the distribution of PV capacity factors, wind capacity factors, and load demand scenarios from Germany from 2013 to 2015. 
The results show that the PCF can generate realistic PV, wind, and load demand scenarios despite a significant dimensionality reduction, while the FSNF overestimates the probability density function (PDF) in areas of high density and underestimates the tails. 
Moreover, the PDF fits improve with $99\%$ cumulative explained variance (CEV) as opposed to $99.9\%$ CEV, as the training problem becomes better conditioned. 
Besides the PDF, the power spectral density (PSD) analysis reveals that PCF recovers the general periodic behavior of the time series. 
As observed for the PV scenarios, PCA can introduce a potentially desired filter effect. 
On the other hand, there may be a trade-off between accurate distribution fits and inclusion of all the details in the data set. 
Therefore, it is critical to test how PCA-based dimensionality reduction affects the data and the final selection of principal components should always be made considering the desired application of the scenarios. 

In addition to our analysis of the PCF-generated scenarios, we run a comparison to two more established methods, namely Copulas and Wasserstein-GANs. Our results highlight that the PCF performs either similarly good or better in all considered metrics compared to the literature methods. In fact, the PCF is the only method that shows good matches in the PDFs, scores high p-values in the KS-test, does not have any unexpected fluctuation, and recovers the zero variance characteristic of the photovoltaic time series during the nighttime. 

In conclusion, our results highlight the importance of considering the inherent latent dimensionality of the data when using normalizing flow generative models, and that the combination of PCA dimensionality reduction and normalizing flows can build powerful distribution models for energy time series scenarios. 

%% file: sections/acknowledgement.tex



\section*{Acknowledgements}
\label{sec:acknowledgements}
This work was performed as part of the Helmholtz School for Data Science in Life, Earth and Energy (HDS-LEE) and received funding from the Helmholtz Association of German Research Centres.
This publication is based upon work supported by the King Abdullah University of Science and Technology (KAUST) Office of Sponsored Research (OSR) under Award No. OSR-2019-CRG8-4033, and the Alexander von Humboldt Foundation.\\
We thank the anonymous reviewers for their valuable feedback which greatly helped us to improve our manuscript.

%% file: sections/appendices.tex

\appendix
\section*{Nomencalture}
\label{sec:appendix}
\subsection*{Acronyms}
\noindent\begin{tabularx}{\columnwidth}{lX}
        ANN  & Artificial neural network      \\
        CEV  & Cumulative explained variance \\
        CVF  & Change of variables formula      \\
        DGM  & Deep generative model      \\
        FSNF & Full space normalizing flow    \\
        GAN  & Generative adversarial network   \\
        KDE  & Kernel density estimation      \\
        NICE & Nonlinear independent component estimation      \\
        PV   & Photovoltaic      \\
        PSD  & Power spectral density      \\
        PCA  & Principal component analysis      \\
        PCF  & Principal component flow      \\
        PDF  & Probability density function      \\
        realNVP  & Real non-volume preserving transformation      \\
        VAE  & Variational autoencoder      \\
\end{tabularx}

\subsection*{Greek letters}
\noindent\begin{tabularx}{\columnwidth}{lX}
        $\theta$            & Trainable parameters of conditioner networks $s_\theta$ and $t_\theta$       \\
        $\mathbf{\mu}_X$    & Mean value of distribution $X$             \\
        $\psi$              & Injective transformation      \\
\end{tabularx}

\subsection*{Latin letters}
\noindent\begin{tabularx}{\columnwidth}{lX}
        $d$                 & Dimension                     \\
        $D$                 & Dimensionality of target distribution          \\
        $f$                 & Transformation                \\
        $f_{CL}$            & Coupling layer transformation \\
        $\mathbf{I}$        & Identity matrix               \\
        $\mathbf{J}$        & Jacobian                      \\
        $K$                 & Number of transformations in composition \\ 
        $M$                 & Dimensionality $<D$                \\
        $p_X/p_Y$           & Probability density of $X$ and $Y$\\
        $s_\theta$          & Conditioner network           \\
        $t_\theta$          & Conditioner network           \\
        $\mathbf{V}$        & Matrix of right singular vectors \\        
        $\mathbf{V}_P$      & Truncated matrix of right singular vectors \\
        $\mathbf{x}$        & Target distribution sample    \\
        $\mathbf{\tilde{x}}$& Compressed target distribution sample   \\
        $X$                 & Target distribution           \\
        $\mathbf{z}$        & Base distribution sample      \\
        $Z$                 & Base distribution             \\
\end{tabularx}